\documentclass{article}

\usepackage{PRIMEarxiv}

\usepackage[utf8]{inputenc} 
\usepackage[T1]{fontenc}    
\usepackage{hyperref}       
\usepackage{url}            
\usepackage{booktabs}       
\usepackage{amsfonts}       
\usepackage{nicefrac}       
\usepackage{microtype}      
\usepackage{lipsum}
\usepackage{fancyhdr}       
\usepackage{graphicx}       
\graphicspath{{media/}}     

\title{AdaptaGen: Domain-Specific Image Generation through Hierarchical Semantic Optimization Framework
}

\author{
  Suoxiang Zhang \\
  China Agricultural University \\
  Beijing, China\\
  \texttt{gravityzh@cau.edu.cn} \\
   \And
   Xiaxi Li \\
   University of New South Wales \\
   Sydney, Australia \\
   \And  
  Hongrui Chang \\
  China Agricultural University \\
  Beijing, China\\
  \And  
  Zhuoyan Hou \\
  China Agricultural University \\
  Beijing, China\\
  \And  
  Guoxin Wu \\
  China Agricultural University \\
  Beijing, China\\
  \And  
  Ronghua Ji \\
  China Agricultural University \\
  Beijing, China\\
  \texttt{jessic1212@cau.edu.cn} \\
}

\begin{document}
\maketitle

\begin{abstract}
Domain-specific image generation aims to produce high-quality visual content for specialized fields while ensuring semantic accuracy and detail fidelity. However, existing methods exhibit two critical limitations: First, current approaches address prompt engineering and model adaptation separately, overlooking the inherent dependence between semantic understanding and visual representation in specialized domains. Second, these techniques inadequately incorporate domain-specific semantic constraints during content synthesis, resulting in generation outcomes that exhibit hallucinations and semantic deviations. To tackle these issues, we propose AdaptaGen, a hierarchical semantic optimization framework that integrates matrix-based prompt optimization with multi-perspective understanding, capturing comprehensive semantic relationships from both global and local perspectives. To mitigate hallucinations in specialized domains, we design a cross-modal adaptation mechanism, which, when combined with intelligent content synthesis, enables preserving core thematic elements while incorporating diverse details across images. Additionally, we introduce a two-phase caption semantic transformation during the generation phase. This approach maintains semantic coherence while enhancing visual diversity, ensuring the generated images adhere to domain-specific constraints. Experimental results confirm our approach's effectiveness, with our framework achieving superior performance across 40 categories from diverse datasets using only 16 images per category, demonstrating significant improvements in image quality, diversity, and semantic consistency.
\end{abstract}

\keywords{Domain-Specific Image Generation, Prompt Engineering, Low-Rank Adaptation (LoRA)}

\begin{figure}
  \includegraphics[width=\textwidth]{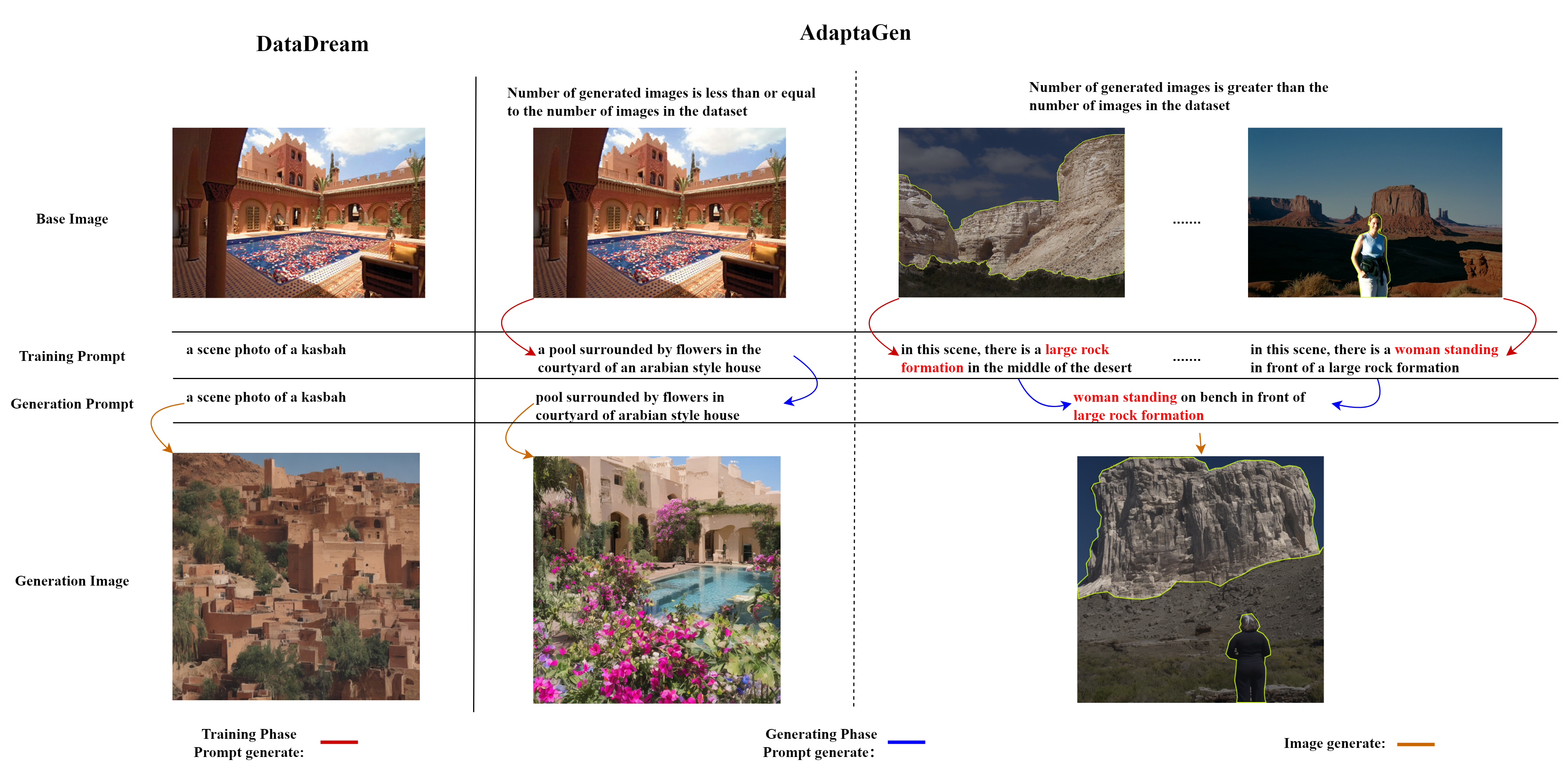}
  \caption{Comparison of prompt handling strategies: AdaptaGen's semantic transformation approach versus DataDream's template-based method for image generation.}
  \label{fig:teaser}
\end{figure}

\section{Introduction}
In many specialized fields requiring domain expertise, data collection often requires significant investments of human, material, and time resources \cite{birhane2021multimodal}. The complexity and diversity of environmental factors further add to the challenges of data acquisition, which limits the scope of many studies due to data scarcity \cite{fu2024ap}. However, with the rapid development and maturation of generative model technologies, new solutions have emerged for these fields \cite{rombach2022high, ramesh2021zero}.

Through generative models, we can not only simulate data from real-world scenarios but also generate diverse, high-quality data tailored to specific needs, thus supporting both research and practical applications \cite{liu2024dora, zhang2023adalora}. This approach not only reduces the cost of obtaining specialized data but can, in some cases, bridge data gaps and maintain data balance \cite{dettmers2023qlora}.

However, despite the tremendous potential demonstrated by generative models in enriching specialized data, their performance remains highly dependent on the quality of prompt design \cite{liu2022design, oppenlaender2022taxonomy}. Currently, most text-to-image models still show insufficient emphasis on prompt processing, primarily relying on preset templates to generate relatively simple prompts \cite{wang2022diffusiondb}. While this approach enables rapid batch generation, it often neglects specific entities and detailed features across different scenarios, leading not only to misguided image generation but also frequently resulting in duplicate images during large-scale generation, ultimately producing outputs with evident biases and distortions \cite{bianchi2023easily}.

This simplified prompting strategy is particularly problematic in specialized domains such as agriculture and medicine, which contain numerous professional terms and concepts. Without appropriate contextual understanding, models struggle to correctly interpret specialized terminology, leading to semantic deviations and hallucinations \cite{xu2024prompt}. As illustrated in Figure~\ref{fig:teaser}, our semantic transformation approach addresses these limitations by preserving core thematic elements while enabling controlled variation, in contrast to template-based methods that often neglect specific entities and detailed features across different scenarios. These hallucinations not only affect the accuracy of generated content but may also produce misleading results in practical applications, especially in research and diagnostic environments where precision is critical for decision-making and knowledge advancement~\cite{pavlichenko2023genetic}.

These challenges highlight a fundamental issue in domain-specific image generation: how to accurately capture domain-specific semantic features while reducing model hallucinations in data-scarce scenarios. Addressing this issue requires a comprehensive approach that focuses on two core aspects: (1) integrating prompt optimization and model adaptation to leverage the inherent dependence between semantic understanding and visual representation in specialized domains, and (2) incorporating domain-specific semantic constraints during content synthesis to mitigate hallucinations and semantic deviations.

While identifying these core aspects provides a theoretical foundation for our approach, implementing effective solutions introduces additional practical challenges. In specialized domains, obtaining large amounts of high-quality training data is often difficult and costly, frequently limiting researchers to working with very small sample sizes. This data scarcity directly impacts our ability to address the identified issues, as traditional methods typically rely on abundant domain-specific examples. Full fine-tuning of large pre-trained models is not only computationally expensive but also prone to overfitting when data is limited, further exacerbating hallucination issues. Therefore, to effectively implement our comprehensive approach, we need to explore techniques that can efficiently adapt to specific domains with limited data while maintaining semantic accuracy and reducing hallucinations.

Compared to traditional full model fine-tuning, Low-Rank Adaptation (LoRA) techniques approximate weight updates through low-rank matrix decomposition, enabling efficient domain adaptation by adjusting only a small number of parameters \cite{hu2022lora}. Traditional fine-tuning requires updating all parameters of a model, which is not only computationally expensive but also requires substantial domain data to avoid overfitting. In contrast, LoRA keeps the original model weights frozen while adding trainable small matrices for adjustments. This lightweight approach is particularly suitable for domain-specific image generation in resource-constrained environments, where both computational resources and domain-specific data are limited \cite{kopiczko2024vera}.

Recent research has demonstrated that parameter-efficient fine-tuning methods like LoRA exhibit heightened sensitivity to prompt details compared to full model fine-tuning \cite{hao2023optimizing}. This sensitivity arises because LoRA adaptations specifically target cross-attention mechanisms that align textual prompts with visual content, creating a more direct pathway between prompt semantics and generated visual features \cite{kim2024datadream}. In domain-specific image generation tasks, this prompt sensitivity presents both a challenge and an opportunity for achieving more precise control over generated content.

Given this heightened sensitivity to prompts in LoRA-based approaches, the quality and precision of input prompts become increasingly critical for successful domain adaptation. Our goal is to generate more precise and semantically rich prompts for each image, thereby mitigating bias from the root and guiding the model to more comprehensively reproduce the information within the image \cite{cao2023beautifulprompt}. Specifically, more accurate prompts can better capture the details and semantic relationships within a scene, improving the realism and consistency of the generated images \cite{rosenman2024neuroprompts}.

While prompt quality is essential, relying entirely on manually generated prompts introduces several significant challenges in practice. First, human descriptions may lack objectivity, being influenced by personal understanding or cultural background, which could result in hidden or explicit biases in the descriptions \cite{bianchi2023easily}. Second, manually created descriptions often fail to cover all essential information of an image, potentially leading to incomplete or incorrect guidance, thus further affecting the quality of the generated results \cite{wang2024promptcharm}. Finally, large-scale generation requires substantial human and financial resources, making manual prompt creation impractical for many applications. Therefore, developing an approach to automatically generate precise prompts is crucial, as it reduces the issues caused by human intervention and significantly enhances the applicability and performance of text-to-image models in diverse scenarios \cite{he2024automated}.

In this paper, we propose an automated prompt optimization and domain image generation framework, AdaptaGen, focusing on addressing the accuracy, hallucination, and performance challenges faced by generative models in diverse scenarios within fixed domains. Our framework effectively mitigates hallucination issues in specialized domain generation through innovative hierarchical semantic learning and feature fusion mechanisms, requiring as few as 16 images per category to achieve effective domain adaptation and significantly reducing semantic deviations and misinterpretations.

The AdaptaGen framework operates in a structured three-stage process. First, our matrix-based prompt optimization stage uses BLIP-2 \cite{li2023blip} to analyze each input image from multiple perspectives, generating diverse candidate descriptions, followed by CLIP-based \cite{radford2021learning} semantic alignment to rank and select optimal descriptions. This multi-perspective understanding ensures accurate capture of domain-specific entities and relationships, reducing the risk of semantic deviation and model hallucination. Second, the cross-modal adaptation stage employs LoRA fine-tuning with carefully controlled parameter updates, focusing on key attention mechanisms to learn domain-specific features while maintaining model stability. Finally, we propose a two-phase caption semantic transformation during the generation phase using the T5 model \cite{raffel2020exploring}: initially summarizing original captions with controlled randomness to preserve core features while enabling variations, then combining captions into a corpus when additional prompts are needed, extracting and fusing elements across images. This approach maintains semantic coherence while enhancing visual diversity, ensuring the generated images adhere to domain-specific constraints. 

Our contributions can be summarized as follows:

\textbf{(1) Integrated Semantic-Visual Framework for Domain-Specific Generation}: We propose a novel framework that explicitly addresses the inherent dependence between semantic understanding and visual representation in specialized domains. By integrating matrix-based prompt optimization with cross-modal adaptation, our approach treats prompt engineering and model adaptation as interdependent components within a unified optimization process, rather than separate stages. This integration enables automated prompt ranking and semantic alignment through structured learning, significantly reducing manual intervention while ensuring domain-appropriate representations, thereby improving both generation quality and semantic accuracy.

\textbf{(2) Domain-Specific Semantic Constraint Incorporation through Advanced Transformation}: We introduce a two-phase caption semantic transformation mechanism that effectively incorporates domain-specific semantic constraints during content synthesis. This mechanism preserves core thematic elements while enabling controlled variation, significantly reducing hallucinations and semantic deviations in specialized domains. By maintaining semantic boundaries during the generation process, our approach ensures that generated images adhere to domain-specific constraints while still allowing for visual diversity.

\textbf{(3) Comprehensive Hallucination Mitigation for Resource-Constrained Specialized Domains}: Our framework integrates the above innovations to address the challenge of generating accurate content in data-scarce specialized domains. The hierarchical semantic understanding mechanism accurately captures domain-specific terminology and relationships from as few as 16 images per category, while the intelligent content synthesis ensures detail fidelity in generated outputs. This comprehensive approach enables high-quality image generation in specialized domains with extremely limited training data, addressing a critical gap in current generative approaches. Experimental results demonstrate significant improvements in both distribution alignment and semantic consistency compared to current representative approaches across 40 categories from diverse datasets.

Extensive experimental results demonstrate the effectiveness of our approach. On a diverse set of 40 randomly selected categories from ImageNet100 \cite{deng2009imagenet}, Sun397 \cite{xiao2010sun}, Food101 \cite{bossard2014food}, and PlantLeafDiseases \cite{geetharamani2019identification} datasets, our method achieves superior performance in multiple dimensions: distribution alignment, image quality and diversity, and semantic consistency. Using only 16 images per category, our framework demonstrates significant improvements over current state-of-the-art methods, validating the effectiveness of AdaptaGen in domain-specific image generation, particularly in scenarios requiring precise semantic control and fine-grained detail preservation with extremely limited training data.

\section{Related Work}

\subsection{Prompt Engineering}

Prompt engineering is fundamental to text-to-image (T2I) generation, as diffusion models like Stable Diffusion \cite{rombach2022stable} and DALL-E \cite{ramesh2021dalle} critically depend on prompt quality. Research demonstrates that conventional textual descriptions often fail to capture desired visual details \cite{xu2024prompt}, presenting significant challenges in specialized domains.

Effective prompt crafting requires domain expertise that many users lack \cite{liu2022design, oppenlaender2022taxonomy}, while representation bias remains a persistent issue. Studies show T2I models amplify demographic stereotypes \cite{bianchi2023easily} and perpetuate social biases from training datasets \cite{birhane2021multimodal}. Despite various optimization approaches—including reinforcement learning techniques \cite{cao2023beautifulprompt, hao2023optimizing}, large language model refinement \cite{he2024automated}, genetic algorithms \cite{pavlichenko2023genetic}, and mixed-initiative systems \cite{wang2024promptcharm}—most T2I models still rely on preset templates generating relatively simple prompts \cite{wang2022diffusiondb, kim2024datadream}, neglecting specific entities and detailed features crucial for specialized domains.

These approaches typically treat prompt engineering separately from model adaptation, overlooking their interdependence. This separation is particularly problematic in data-scarce specialized domains where limited examples make it difficult to establish semantic-visual relationships, often resulting in hallucinations when interpreting specialized terminology.

\subsection{LoRA and Parameter-Efficient Fine-Tuning}

Parameter-Efficient Fine-Tuning (PEFT) methods reduce the computational burden of adapting large pre-trained models through adapter-based methods \cite{houlsby2019parameter, he2021towards}, prompt-based methods \cite{li2021prefix, lester2021power}, and weight modification methods \cite{hu2022lora, zaken2021bitfit}. LoRA (Low-Rank Adaptation) \cite{hu2022lora} has gained particular attention for approximating weight updates through low-rank decomposition while preserving original model weights—ideal for domain-specific generation with limited datasets.

Several variants enhance LoRA's capabilities: AdaLoRA \cite{zhang2023adalora} dynamically allocates parameters based on importance; QLoRA \cite{dettmers2023qlora} combines low-rank adaptation with quantization; DoRA \cite{liu2024dora} decomposes weights into magnitude and direction components; and VeRA \cite{kopiczko2024vera} reduces parameters by sharing random matrices across layers. However, these approaches primarily focus on model adaptation without addressing the entire generation pipeline.

Current LoRA-based methods inadequately incorporate domain-specific semantic constraints during content synthesis, especially with extremely limited data. They typically adjust model parameters without sufficient regard for domain-specific semantic boundaries, lacking mechanisms for maintaining semantic coherence while enabling visual diversity.

AdaptaGen builds upon LoRA for domain-specific image generation, introducing an adaptive mechanism that controls adjustment intensity to balance pre-trained knowledge and domain-specific features. Unlike previous methods focused solely on model adaptation, our approach addresses the entire generation pipeline from prompt optimization to content synthesis, offering a comprehensive solution for specialized domains with limited training data.

\begin{figure*}[t]
    \centering
    \includegraphics[width=\textwidth]{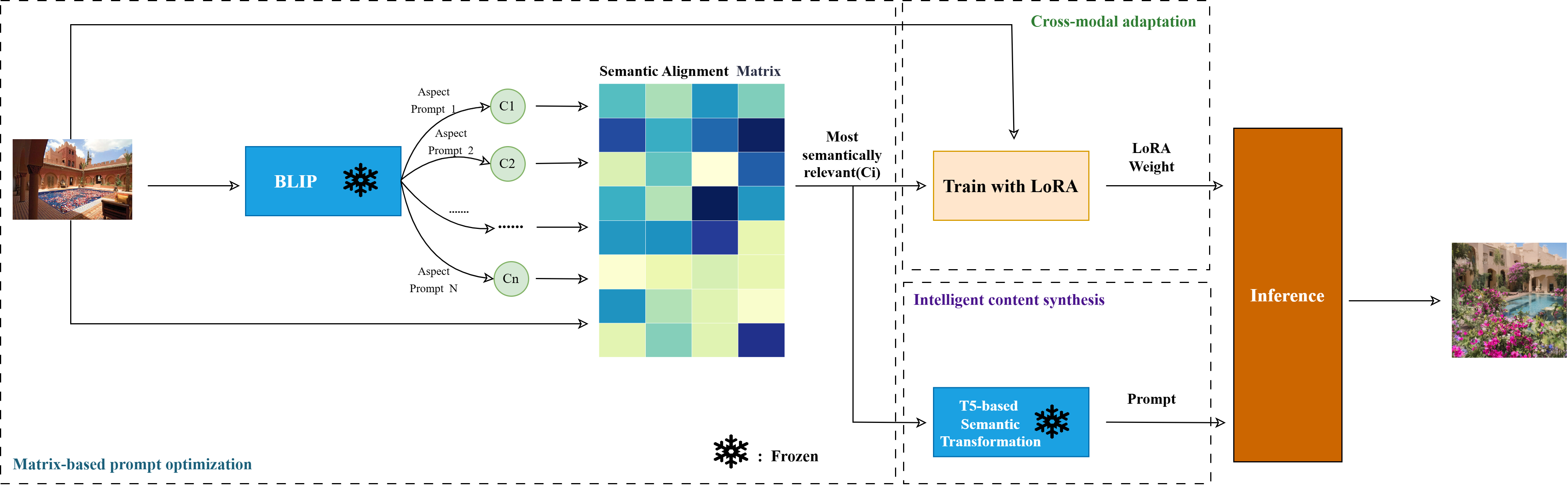}
    \caption{The three-stage hierarchical semantic optimization framework of AdaptaGen, illustrating matrix-based prompt optimization, cross-modal adaptation, and intelligent content synthesis processes.}
    \label{fig:model}
\end{figure*}

\section{Method}

In this section, we present AdaptaGen, a novel framework for domain-specific image generation through hierarchical semantic optimization. Our framework can be formally defined as a mapping function $M : X \rightarrow P \rightarrow Y$, where $X$ represents the input image space, $P$ denotes the prompt space, and $Y$ represents the target image space. This mapping is realized through a three-stage process: $M = g \circ f = g(f(x))$, where $f : X \rightarrow P$ represents prompt optimization and $g : P \rightarrow Y$ represents image generation. Figure~\ref{fig:model} provides an overview of our three-stage hierarchical semantic optimization framework, illustrating the flow from matrix-based prompt optimization through cross-modal adaptation to intelligent content synthesis.

\subsection{Matrix-Based Prompt Optimization with Multi-perspective Understanding}

In our framework, matrix-based multi-perspective prompt optimization is key to generating high-quality candidate descriptions. For a batch of input images $\{x_1, x_2, \ldots, x_n\}$, we construct a similarity matrix system to optimize prompt selection.

For each image $x_i$, we use BLIP-2 to generate multiple candidate descriptions: $C_i = \{c_{i1}, c_{i2}, \ldots, c_{im}\}$ through different prompt templates that capture various semantic dimensions including object recognition, scene composition, subject emphasis, and contextual interpretation. For the entire dataset, we obtain:

\begin{equation}
C = \bigcup_{i=1}^{n} C_i
\end{equation}

The core of our approach is constructing an image-description similarity matrix $S$, where each element $S(i,j)$ represents the semantic similarity between image $x_i$ and description $c_{ij}$:

\begin{equation}
S(i,j) = \frac{(E_I(x_i))^T E_T(c_{ij})}{\|E_I(x_i)\|\|E_T(c_{ij})\|}
\end{equation}

where $E_I(x_i)$ and $E_T(c_{ij})$ represent image and text embeddings from CLIP's encoders. Through this matrix optimization, we select the most semantically relevant description for each image:

\begin{equation}
c^*_i = \arg\max_{j\in\{1,\ldots,m\}} S(i,j)
\end{equation}

This approach efficiently generates optimal descriptions for the entire dataset in batch processing, significantly reducing semantic deviations common in specialized domains. By capturing multiple perspectives, our method accurately identifies domain-specific entities and relationships, effectively mitigating potential hallucinations in subsequent generation steps. This is particularly crucial when working with limited training data, as each image must contribute maximum semantic information.

\subsection{Cross-Modal Adaptation}

To adapt pretrained models to specialized domains with minimal training data, we employ Low-Rank Adaptation (LoRA) guided by a diffusion process loss function:

\begin{equation}
L_{diff}(\theta) = \mathbb{E}_{x,t,\epsilon} \left[ \|\epsilon - \epsilon_\theta(x_t, t, E(c^*))\|^2 \right]
\end{equation}

where $\epsilon_\theta$ is the noise prediction network, $x_t$ is the noised image at timestep $t$, and $E(c^*)$ is the embedding of the optimized caption. 

Unlike traditional approaches that treat model adaptation separately from prompt engineering, our method leverages their inherent interdependence by implementing LoRA's targeted parameter updates:

\begin{equation}
W = W_0 + BA
\end{equation}

where $W_0$ represents the original weight matrix, and $B$, $A$ are low-rank matrices. During inference, we apply a scaling factor to control domain-specific adaptations:

\begin{equation}
h = W_0x + \alpha\frac{r}{d}(BA)x
\end{equation}

This strategic approach focuses updates on key attention mechanisms, ensuring the model learns domain-specific features while preserving the pre-trained model's general capabilities, effectively reducing hallucinations in specialized contexts.

\subsection{Intelligent Content Synthesis through Advanced Semantic Transformation}

Our final component addresses the challenges of limited training data and enables large-scale generation with enhanced diversity through a two-phase T5-based semantic transformation strategy.

When generating fewer images than available captions, our mechanism operates in a controlled randomization mode, where each caption $c^*_i$ undergoes transformation with a dynamic temperature:

\begin{equation}
c'_i = T5(c^*_i, \tau_i)
\end{equation}

where $\tau_i = \tau_{base} + \Delta\tau \cdot u(-1, 1)$ represents a temperature with controlled variability ($\tau_{base}=0.8$, $\Delta\tau=0.2$), and $u(-1, 1)$ is a uniform random variable. This preserves core semantic content while introducing subtle variations.

When generating more images than available captions, we implement an intelligent content synthesis mechanism that:

\begin{enumerate}
    \item Preserves core thematic elements essential for semantic coherence
    \item Achieves cross-image feature fusion, combining elements from multiple sources
    \item Enhances diversity through intelligent recombination of domain-specific attributes
\end{enumerate}

This feature fusion capability is particularly powerful when combined with our LoRA-adapted model. The optimized captions serve as conditional input to the diffusion model:

\begin{equation}
y = DM(c^*, n, \omega, s)
\end{equation}

where $DM$ represents the diffusion model pipeline, $c^*$ is the optimized caption, $n$ is the inference steps, $\omega$ is the guidance scale, and $s$ is the LoRA scale.

Even with just 16 images per category, our intelligent content synthesis mechanism can generate hundreds of semantically accurate and visually diverse images while maintaining domain-specific characteristics and significantly reducing hallucinations compared to existing approaches. Experimental results demonstrate a substantial improvement in semantic consistency as evidenced by higher CLIP scores across all domains, indicating our framework's enhanced ability to mitigate, though not completely eliminate, various types of semantic deviations common in specialized generation tasks.

In summary, our three-stage AdaptaGen framework effectively addresses domain-specific image generation challenges through: (1) a hierarchical semantic optimization framework integrating prompt optimization and model adaptation, and (2) an intelligent content synthesis mechanism incorporating domain-specific semantic constraints. This approach offers a comprehensive solution for generating high-quality, diverse images in resource-constrained specialized domains, as validated by our experimental results across multiple metrics and diverse categories.

\section{Experiments}

We conduct extensive experiments to evaluate the effectiveness of AdaptaGen across diverse domains. We first describe our experimental setup, followed by a comprehensive analysis of results with particular emphasis on our framework's ability to address the two core challenges in domain-specific image generation: (1) integrating prompt optimization and model adaptation, and (2) incorporating domain-specific semantic constraints during content synthesis.

\subsection{Experimental Setup}

\paragraph{Datasets.}
To comprehensively evaluate our approach, we select four representative datasets with different characteristics: (1) \textbf{Food101}~\cite{bossard2014food}, containing food categories requiring precise texture representation; (2) \textbf{ImageNet100}~\cite{deng2009imagenet}, a subset of ImageNet with diverse object categories; (3) \textbf{Sun397}~\cite{xiao2010sun}, featuring scene categories with complex spatial relationships; and (4) \textbf{PlantLeafDiseases (PLDiseases)}~\cite{geetharamani2019identification}, representing specialized domain applications. From each dataset, we randomly select 10 categories, for a total of 40 categories. Importantly, we extract only 16 images from each category as the training dataset, deliberately creating an extremely data-scarce scenario to test our framework's effectiveness in resource-constrained environments.

\paragraph{Baselines.}
We compare AdaptaGen with two representative methods from recent literature: (1) \textbf{DoRA}~\cite{liu2024dora}, a method that decomposes neural network weights into magnitude and direction components; and (2) \textbf{DataDream}~\cite{kim2024datadream}, which creates synthetic datasets from limited samples using optimized prompt templates. All methods use Stable Diffusion v2.1 as the base model for fair comparison and are trained using the same 16-image-per-category dataset.

\paragraph{Evaluation Metrics.}
We employ three complementary metrics to evaluate different aspects of generation performance:
\begin{itemize}
    \item \textbf{Fréchet Inception Distance (FID-1k)}: We generate 1,000 images per category to assess distribution alignment between generated and real images.
    \item \textbf{Inception Score (IS)}: We evaluate both quality and diversity of generated images.
    \item \textbf{CLIP Score}: We measure semantic alignment between generated images and their corresponding prompts.
\end{itemize}

These metrics together provide a comprehensive evaluation of our framework's ability to address the challenges of domain-specific image generation. FID assesses distribution alignment, directly reflecting how well our integration of prompt optimization and model adaptation captures domain-specific visual features. IS measures quality and diversity, revealing the effectiveness of our semantic transformation in preventing hallucinations while maintaining diversity. CLIP Score evaluates semantic consistency, demonstrating how well our approach incorporates domain-specific semantic constraints during synthesis.

\subsection{Comparative Performance Analysis}

\subsubsection{Distribution Alignment Analysis}

The FID-1k scores across all 40 evaluation categories are summarized in Table~\ref{tab:avg_fid_comparison}. Our method outperforms baseline approaches in the majority of categories despite being trained on only 16 images per category. As shown in the table, AdaptaGen achieves average FID-1k scores of 65.04 on Food101, 68.90 on ImageNet100, 119.56 on Sun397, and 70.08 on PlantLeafDiseases. Overall, our method achieves an average FID-1k score of 80.90 across all categories, compared to 156.80 for DoRA and 109.64 for DataDream.

Examining specific categories across datasets, our approach shows substantial improvements in fine-grained food categories from Food101. For complex biological subjects in ImageNet100, we observe consistent improvements particularly in categories with intricate structural features. In scene understanding from Sun397, categories with complex spatial arrangements show significant FID improvements, demonstrating our method's capability in handling complex environments with multiple elements and spatial relationships.

The variant of our approach without the feature fusion mechanism (Ours(w/o)) achieves an overall FID-1k score of 126.64, which is still better than DoRA but falls behind our complete method, highlighting the importance of this component in effectively integrating prompt optimization with model adaptation. These results clearly demonstrate that our approach to treating prompt engineering and model adaptation as interdependent components yields superior distribution alignment compared to methods that address these aspects separately.

\begin{table}[t]
\centering
\caption{Average FID-1k scores across different datasets ($\downarrow$)}
\label{tab:avg_fid_comparison}
\setlength{\tabcolsep}{4pt}
\small
\renewcommand{\arraystretch}{1.2}
\begin{tabular}{lccccc}
\toprule
Method & Food101 & ImageNet & Sun397 & PLDiseases & Overall \\
\midrule
Dora & 100.72 & 104.06 & 151.99 & 270.43 & 156.8 \\
DataDream & 100.29 & 93.3 & 159.75 & 85.21 & 109.64 \\
Ours(w/o) & 92.03 & 105.32 & 164.56 & 144.67 & 126.64 \\
Ours & \textbf{65.04} & \textbf{68.9} & \textbf{119.56} & \textbf{70.08} & \textbf{80.9} \\
\bottomrule
\end{tabular}
\end{table}

\subsubsection{Quality and Diversity Evaluation}

Table~\ref{tab:is_comparison} presents Inception Scores across the four datasets. Our method achieves competitive or superior IS values compared to baseline methods on most datasets, demonstrating its ability to maintain high quality and diversity even with extremely limited training data.

On ImageNet100, our method achieves an IS of 2.83, which significantly outperforms both baseline methods (Dora: 2.26, DataDream: 1.78). Notably, when viewed alongside our superior FID score (68.9 compared to Dora's 104.06 and DataDream's 93.3) and higher CLIP score (0.32 versus 0.30), this demonstrates that our approach not only generates more distinctive features but also maintains better distribution alignment and semantic consistency. The fact that our IS exceeds the original dataset's score (1.56) while simultaneously achieving better distribution alignment suggests that our generative process effectively captures the essential characteristics of these natural object categories while reducing noise and redundancy present in the limited training samples.

Notably, while DoRA shows higher IS on Food101 (3.35 vs. our 2.38), the FID scores indicate better distribution alignment with our method, and our IS score (2.38) is closer to the original dataset (2.43). This suggests that our approach more accurately captures the true distribution characteristics of food categories while maintaining image quality.

The variant without the feature fusion mechanism (Ours(w/o)) shows lower IS scores across all datasets compared to our complete method, further confirming the importance of the feature fusion mechanism in enhancing image quality and diversity. These results validate the effectiveness of our approach in incorporating domain-specific semantic constraints during content synthesis, resulting in generated images that exhibit high quality and diversity while minimizing hallucinations.

\begin{table}[t]
\centering
\caption{Inception Score comparison across different datasets ($\uparrow$)}
\label{tab:is_comparison}
\setlength{\tabcolsep}{4pt}
\small
\renewcommand{\arraystretch}{1.2} 
\begin{tabular}{lccccc}
\toprule
Method & Sun397 & PLDiseases & Food101 & ImageNet & Overall \\
\midrule
Dora & 2.93 & 2.97 & \textbf{3.35} & 2.26 & 2.28 \\
DataDream & 2.26 & 2.52 & 1.98 & 1.78 & 2.14 \\
Ours(w/o) & 2.05 & 2.18 & 2.12 & 1.94 & 2.18 \\
Ours & 3.12 & 3.13 & 2.38 & \textbf{2.83} & \textbf{2.87} \\
Original dataset & \textbf{3.26} & \textbf{3.28} & 2.43 & 1.56 & 2.63 \\
\bottomrule
\end{tabular}
\end{table}

\subsubsection{Semantic Consistency Analysis}

Table~\ref{tab:clip_comparison} presents CLIP scores across the datasets, measuring the semantic alignment between generated images and their prompts. These scores are particularly relevant to evaluating how effectively our approach incorporates domain-specific semantic constraints during content synthesis.

Our method consistently achieves the highest CLIP scores across all datasets, with particularly notable improvements in PlantLeafDiseases (0.32 compared to DoRA's 0.28). This indicates stronger semantic alignment between generated images and their text descriptions, which is critical for specialized domains where precise semantic understanding is important.

On Sun397, our method achieves a CLIP score of 0.31 (compared to DoRA's 0.29 and DataDream's 0.28), indicating better semantic alignment in complex scene understanding. On Food101 and ImageNet100 datasets, our method achieves CLIP scores of 0.31 and 0.32 respectively, both outperforming the baseline methods.

Overall, our method achieves an average CLIP score of 0.32 across all four datasets, showing significant improvement over DoRA and DataDream's 0.29. This result further confirms the effectiveness of our hierarchical semantic optimization framework in capturing domain-specific semantic features, as well as validating our content synthesis mechanism's ability to maintain semantic coherence while enhancing visual diversity.

\begin{table}[t]
\centering
\caption{CLIP Score comparison across different datasets($\uparrow$)}
\label{tab:clip_comparison}
\setlength{\tabcolsep}{4pt}
\small
\renewcommand{\arraystretch}{1.2}
\begin{tabular}{lccccc}
\toprule
Method & Sun397 & PLDiseases & Food101 & ImageNet & Overall \\
\midrule
Dora & 0.29 & 0.28 & 0.29 & 0.30 & 0.29 \\
DataDream & 0.28 & 0.27 & 0.29 & 0.30 & 0.29 \\
Ours & \textbf{0.31} & \textbf{0.32} & \textbf{0.31} & \textbf{0.32} & \textbf{0.32} \\
\bottomrule
\end{tabular}
\end{table}

\subsubsection{Ablation Study}

To evaluate the contribution of the two-phase caption semantic transformation mechanism, we compare the complete AdaptaGen (Ours) with a variant without this component (Ours(w/o)). As shown in Table~\ref{tab:avg_fid_comparison}, removing the semantic transformation mechanism increases the average FID-1k score from 80.90 to 126.64 across all datasets. Additionally, as shown in Table~\ref{tab:is_comparison}, the IS score decreases from 2.87 to 2.18.

Analysis of our ablation study reveals that the caption semantic transformation provides the most significant improvements in categories with complex compositional elements in Food101 and categories requiring specialized domain knowledge in PlantLeafDiseases. This demonstrates the mechanism's effectiveness in preserving core thematic elements while incorporating diverse details across images—a key capability for domain-specific generation.

The controlled randomization in the first phase of our caption semantic transformation preserves semantic coherence while enabling sufficient variation, while the cross-image feature fusion in the second phase effectively combines elements from multiple source images to form novel yet semantically accurate descriptions. Without this mechanism, the generated images show less diversity and more frequent hallucinations, particularly in domains with complex technical terminology and visual attributes. These results confirm that our two-phase caption semantic transformation is crucial for maintaining the balance between semantic accuracy and visual diversity in domain-specific image generation, especially when working with extremely limited training data.

\subsection{Domain-Specific Performance and Hallucination Reduction}

Our experiments evaluate AdaptaGen's effectiveness in addressing domain-specific image generation challenges, with particular focus on hallucination mitigation across diverse specialized domains.

\subsubsection{Multi-Metric Performance Across Domains}

Tables~\ref{tab:avg_fid_comparison},~\ref{tab:is_comparison}, and~\ref{tab:clip_comparison} present the performance of AdaptaGen and baseline methods across the evaluation datasets. The integrated analysis of these metrics reveals several key findings regarding hallucination mitigation in domain-specific generation.

\paragraph{Specialized Domain Hallucination Reduction.} The most significant improvements are observed in the PlantLeafDiseases dataset, where domain expertise and precise semantic control are critical. Our method achieves an FID-1k score of 70.08 compared to DoRA's 270.43, while simultaneously maintaining high IS (3.13 vs. 2.97) and superior CLIP scores (0.32 vs. 0.28). This combined improvement across all three metrics demonstrates our framework's capacity to capture specialized visual features while maintaining semantic accuracy—the core challenge in reducing hallucinations in specialized domains.

\paragraph{Cross-Modal Alignment in Complex Scenes.} In the Sun397 dataset, featuring complex scenes with numerous spatial relationships, our method demonstrates strong performance across all metrics. The combined improvement in FID and CLIP scores is particularly relevant to our goal of reducing spatial relationship hallucinations through improved cross-modal alignment. Across our evaluation, categories with complex spatial arrangements show substantial improvements, indicating that our hierarchical semantic understanding framework effectively captures the functional relationships between scene elements—a key factor in reducing spatial hallucinations.

\paragraph{Fine-Grained Detail Preservation.} In Food101, where precise texture and composition details are essential, our method achieves the lowest FID score (65.04) while maintaining competitive IS (2.38) and the highest CLIP score (0.31). The combination of improved distribution alignment and semantic consistency metrics demonstrates our framework's ability to reduce attribute hallucinations by preserving fine-grained details—addressing another key challenge in domain-specific generation.

\paragraph{Structural Accuracy in Biological Subjects.} For biological subjects in ImageNet100, our method demonstrates remarkable performance across all metrics (FID: 68.9, IS: 2.83, CLIP: 0.32). The exceptionally high IS—surpassing even the original dataset (1.56)—combined with the lowest FID and highest CLIP scores indicates that AdaptaGen successfully captures structural correctness while maintaining diversity. This addresses the structural hallucination problem that commonly affects biological subject generation.

\subsubsection{Component Contribution to Hallucination Reduction}

Our ablation study results, comparing AdaptaGen with and without the feature fusion mechanism, provide insights into how different components contribute to hallucination mitigation.

The feature fusion mechanism substantially improves performance across all datasets, reducing the overall FID-1k score from 126.64 to 80.9 and increasing the IS score from 2.18 to 2.87. This improvement is particularly pronounced in specialized domains and categories with complex feature relationships. Our analysis indicates that the feature fusion mechanism provides the most significant improvements in categories requiring specialized domain knowledge, especially in the PlantLeafDiseases dataset. This demonstrates the mechanism's crucial role in capturing domain-specific visual patterns accurately while maintaining semantic coherence.

The correlation between FID improvements, IS enhancements, and CLIP score increases across datasets suggests that the fusion mechanism contributes significantly to both visual quality and semantic consistency. This aligns with our framework's goal of reducing hallucinations through improved semantic understanding and cross-modal alignment, validating our approach to incorporating domain-specific semantic constraints during content synthesis.

\subsubsection{Domain-Specific Hallucination Types and Mitigation}

Our multi-metric analysis reveals that AdaptaGen effectively addresses different types of hallucinations across various domains, even when trained on just 16 images per category:

\begin{itemize}
    \item \textbf{Semantic Hallucinations:} The consistent improvement in CLIP scores across all domains, particularly in PlantLeafDiseases (0.32), indicates reduced semantic misalignment between textual descriptions and visual content. This directly addresses the challenge of specialized terminology understanding, especially evident in disease categories where baseline methods fail to capture pathological features despite similar FID scores.
    
    \item \textbf{Attribute Hallucinations:} In Food101, the combined improvement in FID (65.04) and CLIP scores (0.32) demonstrates reduced attribute hallucinations, such as incorrect ingredient combinations or implausible food compositions. The consistently improved results across food categories confirm our framework's effectiveness in capturing domain-specific attribute relationships.
    
    \item \textbf{Spatial Relationship Hallucinations:} In complex scene categories from Sun397, our method's improved FID (119.56) and CLIP scores (0.31) reflect reduced spatial relationship hallucinations. The significant improvements in categories with complex environmental layouts demonstrate how our hierarchical semantic understanding approach captures functional relationships between objects in complex environments.
    
    \item \textbf{Structural Hallucinations:} In ImageNet100 biological categories, the exceptional performance across all metrics, particularly IS (2.83), indicates reduced structural hallucinations. Our framework successfully captures species-specific features and proportions across various biological subjects.
\end{itemize}

\subsubsection{Summary of Findings}

The integrated analysis of FID, IS, and CLIP metrics demonstrates that AdaptaGen effectively addresses the key challenges in domain-specific image generation through its hierarchical semantic optimization framework. The consistent improvement across all three metrics, particularly in specialized domains and complex categories, confirms our approach's effectiveness in reducing various types of hallucinations while maintaining generation quality and diversity, even when trained on only 16 images per category.

The correlation between improved semantic consistency (CLIP) and distribution alignment (FID) across domains validates our core premise that enhanced semantic understanding leads to reduced hallucinations in domain-specific generation. The framework's strong performance in specialized domains like plant disease diagnosis demonstrates its potential for practical applications in fields requiring domain expertise and precise visual representation.

Our results conclusively demonstrate the effectiveness of (1) integrating prompt optimization and model adaptation to leverage the inherent dependence between semantic understanding and visual representation in specialized domains, and (2) incorporating domain-specific semantic constraints during content synthesis to mitigate hallucinations and semantic deviations. These findings validate our approach as a comprehensive solution for domain-specific image generation in resource-constrained environments.

\section{Conclusion}

In this paper, we presented AdaptaGen, a novel hierarchical semantic optimization framework for domain-specific image generation. By integrating matrix-based prompt optimization with multi-perspective understanding and cross-modal adaptation, AdaptaGen effectively leverages the inherent dependence between semantic understanding and visual representation in specialized domains. Our two-phase caption semantic transformation mechanism further incorporates domain-specific semantic constraints during content synthesis, significantly mitigating hallucinations and semantic deviations.

Our framework achieves superior performance across 40 categories from diverse datasets, demonstrating significant improvements in FID scores (48.4\% over DoRA and 26.2\% over DataDream), while maintaining high image quality and semantic consistency. These comprehensive improvements validate our approach of treating prompt optimization and model adaptation as interdependent components within a unified framework, rather than separate processes.

Particularly noteworthy is our method's effectiveness in specialized domains like plant disease diagnosis, where the accurate capture of domain-specific semantic features and reduction of model hallucinations is crucial. The hierarchical semantic understanding mechanism successfully captures professional terminology and concept relationships, while the intelligent content synthesis ensures detail richness and accuracy in generated content, addressing the key challenges we identified for domain-specific image generation in data-scarce scenarios.

Most significantly, AdaptaGen achieves these results using only 16 images per category, demonstrating its exceptional efficiency in resource-constrained environments. This capability makes our approach particularly valuable for specialized domains where obtaining large training datasets is difficult and costly. Future work will explore extending our framework to handle more complex semantic relationships and incorporate additional domain knowledge sources to further enhance generation fidelity in highly specialized technical fields.

\bibliographystyle{unsrt}  
\bibliography{references}

\end{document}